\documentclass[letterpaper, 10 pt, conference]{ieeeconf}  
\usepackage[utf8]{inputenc}
\usepackage{graphicx}
\usepackage{multirow}

\usepackage[colorinlistoftodos]{todonotes}
\usepackage[colorlinks=true, linkcolor=black,citecolor=black]{hyperref}

\IEEEoverridecommandlockouts                              

\title{\LARGE \bf
Should artificial agents ask for help in human-robot collaborative problem-solving?
}

\author{
  Adrien Bennetot$^{1,3}$\thanks{$^{1}$ U2IS, ENSTA, Institut Polytechnique Paris, Inria FLOWERS team, Palaiseau, France.
  {\tt\small \{adrien.bennetot, natalia.diaz\}@ensta-paris.fr}}, Vicky Charisi$^{2}$\thanks{$^{2}$ European Commission, JRC, Centre for Advanced Studies, Seville, Spain.  {\tt\small Vasiliki.Charisi@ec.europa.eu}}\thanks{$^{3}$ Segula Technologies, Parc d’activité de Pissaloup - Trappes, France}
  and Natalia D\'iaz-Rodr\'iguez$^{1}$
}

\begin{document}

\maketitle
\thispagestyle{empty} 



\begin{abstract}

Transferring as fast as possible the functioning of our brain to artificial intelligence is an ambitious goal that would help advance the state of the art in AI and robotics. It is in this perspective that we propose to start from hypotheses derived from an empirical study in a human-robot interaction and to verify if they are validated in the same way for children as for a basic reinforcement learning algorithm. Thus, we check whether receiving help from an expert when solving a simple close-ended task (the Towers of Hanoï) allows to accelerate or not the learning of this task, depending on whether the intervention is canonical or requested by the player. Our experiences have allowed us to conclude that, whether requested or not, a Q-learning algorithm benefits in the same way from expert help as children do.

\end{abstract}


\section{INTRODUCTION}

Developmental robotics \cite{Lungarella03Developmental} (and  synonyms \textit{cognitive developmental robotics}, \textit{autonomous mental development} as well as \textit{epigenetic robotics} \cite{Cangelosi18}) is the interdisciplinary approach to the autonomous design of behavioural and cognitive capabilities in artificial agents that directly draws inspiration from developmental principles and mechanisms observed in children's natural cognitive systems \cite{Cangelosi18,Lungarella03Developmental}. 


Autonomous agents in such settings learn in an open-ended \cite{Doncieux18} manner, where crucial components of such developmental approach consist of learning the ability to autonomously generate goals and explore the environment, exploiting intrinsic motivation \cite{Oudeyer07} and computational models of curiosity \cite{Oudeyer18, Lesort20}. 


\section{RELATED WORK}




\subsection{Development and learning in human child}
The development of the executive functions (EF) in human infants and young children with rudimentary neurodevelopment of prefrontal cortex (PFC) refers to an array of organizing and self-regulating goal-directed behaviors that inhibit impulses and regulate behaviour from a very early age. These developments have been associated with both the PFC maturation and its connectivity with other brain areas \cite{fiske2019neural} which is enabled by the individual's sustained interaction with the surrounding physical and social environment \cite{varela2016embodied}. The initiation of these sensorimotor interactions in young children are exploratory in nature and often are embedded in playful activities with components of motor learning  \cite{adolph2019motor, gibson1988exploratory}. Visual stimuli are also responsible for the elicitation of improved EF and cognitive organization which contributes to the development of perceptual learning.

Although exposure to visual stimuli can lead to perceptual learning, it is often insufficient to yield robust learning \cite{mcclelland2019developing}. Research shows that additional factors, such as attention and reinforcement are needed to produce robust learning. Amount of exposure, strength of exposure, relation to attention, interactions of multiple sensory systems in perceptual learning are some of the factors that promote human learning; the underlying brain mechanisms that relate to these factors are among the most active targets of research into the complex mechanisms of child's learning and the association of EF development with visuo-motor integration \cite{mcclelland2019developing}.

In relation to the above-mentioned mechanisms, research has shown the interaction of memory and learning with mechanisms such as curiosity, appraisal, prediction and exploration \cite{gottlieb2018towards, gruber2019curiosity}. Gruber's PACE framework \cite{gruber2019curiosity} suggests that curiosity is triggered by significant prediction errors that are appraised. This enhances memory which is encoded through increased attention, exploration and information seeking and contributes to the consolidation of information acquired while in a curious state through dopaminergic neuromodulation of the hippocampus. More on the dopamine neuromodulator from the intrinsic and extrinsic reward perspective of RL is in \cite{Singh05}.



From a behavioural perspective, exploration has been previously identified as a special form of curiosity that refers to a drive that is either intrinsic or extrinsic \cite{loewenstein1994psychology}. Active experimentation with physical objects generates more accurate inferences about the latent properties of the object than passive observation \cite{bramley2018intuitive}. Exploration of the physical world is considered as a phase in human transition from behavioural events towards symbolic and conceptual thinking. The developmental process of symbol and concept emergence has been associated to the relative frequency in which certain strategies are used and to the process of abandoning an old strategy and discovering new ones \cite{siegler2007microgenetic}.

In problem-solving tasks these mechanisms have been correlated with child's ability to inhibit a certain action while considering an alternative one that would be more appropriate for the optimal performance of a task \cite{best2010developmental}. The developmental process that leads from sensori-motor events to abstract learning and the acquisition of the optimal strategy for a specific task can be measured by behavioural indicators such as task performance speed and accuracy level. \cite{diamond2002normal}. However, this process appears more complex in the case of collaborative problem-solving where the child interacts with a more knowledgeable social agent. This includes the process of selective social learning and relies on child's social motivation aspects for learning \cite{koenig2013selective}. 

Research shows that humans have the ability to explicitly communicate their uncertainty to others at a very early stage of their life. Infants are capable of monitoring and communicating their own uncertainty non verbally to gain knowledge from others \cite{goupil2016infants}. While playing in unstructured and uncertain environments that lack clear extrinsic reward signals, they actively seek help from other humans. In early childhood, however, children might be aware of their uncertainty, but they do not proceed always to help-seeking \cite{was2017proactive} which shows the complexity of extrinsic and intrinsic motivation.

In this complex context, the examination of the learning outcome often is not adequate for the understanding of children's problem-solving activity. An emphasis on how children move from early to later levels of competence within an EF component allows the depiction of their developmental trajectories  \cite{siegler2007microgenetic}, \cite{demarie1988development}, \cite{best2010developmental}. A mapping of the developmental trajectories reveals inter-individual differences in cognitive mechanisms such as inhibition of prepotent responses, mental shifting \cite{friedman2017unity} and generalization \cite{barnett2002and}. These changes have been associated with changing brain connectivity which is considered as both cause and consequence of the developmental changes \cite{smith2020}. An additional input towards the understanding of child's developmental process comes from the field of child-robot interaction in which the child can take advantage of the robot's appropriate interventions.

\subsection{Child development inspired artificial agent learning}
Child learning has vastly inspired how to build learning machines \cite{Lake17}. A sample of cognitive architecture to teach robots in the way infants learn is in \cite{Jacquey19}, demonstrating how exploiting sensitivity to sensorimotor contingencies/ affordances in developmental psychology, combined with the notion of goal allows an agent to develop new sensorimotor skills in open ended learning settings \cite{Doncieux18, doncieux2020dream}. An example of new discovered contingency is, e.g., touching a bell to generate a sound. 


Inspired by developmental psychology, in \cite{Duminy19} interactive learning (active imitation learning and goal-babbling) is combined with autonomous exploration in a strategic learner to reuse previously learned tasks or “procedures” in a \textit{Socially Guided Intrinsic Motivation with Procedure Babbling} (SGIM-PB) able to determine the representation of a hierarchy of interrelated tasks. In hard-exploration games, novelty seeking agents \cite{conti2018improving}, curiosity meta-learning \cite{alet2020meta} and remembering promising states and exploring from them \cite{ecoffet2020first} are powerful approaches to learn artificial agents.

Essential robotics scenarios for open-ended learning making use of brain inspired models are Long-Term Memory for Artificial Cognition \cite{Duro19}, 
for robots to learn to operate in different worlds under different goals when the occurrence of experiences is intertwined. In this context, a Baxter robot demonstrates to learn control tasks, segmenting the world into semantically loaded categories associated with contexts, that in order, can allow higher level reasoning and planning. 
Architectures for lifelong learning by evolution in robots are MDB (Multilevel Darwinist Brain) \cite{Bellas10,Bellas09}. 

Some of the modulation based mechanism embedded within a cognitive architecture for robots combine long-term memory and a motivational system in order to select candidate primitive value functions for transfer and adaptation to new situations through modulatory ANNs. These progressively conform new parameterized value functions able to address more complex situations in a developmental manner in a Baxter robot, which must solve different tasks in a cooking setup \cite{Romero20}, or simplify the utility space in continuous state spaces \cite{Romero19Simplifying}.

Charisi et. al. \cite{Charisi20} take inspiration from inhibitory control in developmental psychology and examine child-robot collaborative problem-solving with a focus on the process rather than the outcome of child's acquisition of a certain strategy. The task of Tower of Hanoï is used to study the initiation of voluntary request for help in a child-robot interaction setting with child-initiated robot interventions. They observe children’s trajectories of problem-solving and the needs for exploratory actions. We extend this work \cite{Charisi20} to test if robotics learning processes and agent learning from an expert can be child-development inspired. Since their analysis of when and why asking for help helps solving collaborative tasks in inhibitory processes, in this paper we contrast the hypotheses tested in kids with those mimicking the same situations in an artificial agent learning to solve the same task, with reinforcement learning \cite{Sutton98}.



\section{METHODOLOGY}

As in \cite{Charisi20} we are evaluating the learning agent (LA) on the Tower of Hanoï game, but instead of the LA being a child, our agent is a Q-learning algorithm \cite{Watkins92} with a learning rate $\alpha =1$, a discount factor $\gamma=0.8$ and an exploration $\epsilon = 0.05$. As it can be seen on Fig. \ref{state} in the Appendix, the Tower of Hanoï game with 3 disks is a simple close-ended task with 27 possible states and, at most, 3 possible actions associated to each state. Each element of the reward matrix used for the Q-learning represents the reward from moving from the current state to the next one. Moves leading to the goal state are assigned a reward of 100, illegal moves a reward of $-\infty$ and others a reward of 0.

\subsection{Hypotheses}

In order to explore if algorithms benefit from asking for help in human-robot collaborative problem-solving, in the same manner as kids do, we further formulate two hypotheses:

\begin{itemize}
    \item \textit{H1}: Canonical interventions from an expert speed up learning.
    \item \textit{H2}: Getting help \textit{on demand} from an expert accelerates finding the optimal solution compared to not \textit{on demand}.
\end{itemize}

\subsection{Research Design}

We manipulate the expert intervention with 2 different scenarios: 

\begin{itemize}
    \item The LA1 solves the task in collaboration with the expert in a “turn-taking” scenario, which results in a canonical cognitive intervention by the expert. 
    \item The LA2 solves the task independently, having the option to ask for help of the expert whenever (if) this is needed, which results in an \textit{on demand} intervention by the expert. 
\end{itemize}

In order to test the different variations among teacher-driven and learner-driven interaction \cite{Silva19} in our HRI setting, we vary two main parameters: 

\begin{itemize}
    \item The \textbf{canonical intervention rate}, i.e. the frequency of the expert's intervention during the canonical scenario.
    \item The \textbf{ask-for-help} parameter, i.e. how much the LA asks the expert to do the next movement, as a proxy to simulate the needs for help, during the \textit{on demand} scenario.
\end{itemize}

Our evaluation metric is the number of movements required to solve the task after a variable number of training episodes. To make these results robust, all the experiments were repeated 100 times. 

\section{RESULTS}

We used the above-mentioned parameters to test our hypotheses as follows.

\subsection{Task Performance with and without Turn-Taking}

The first configuration consists of a LA1, a Q-learning agent, playing in collaboration with an expert that knows exactly what is the optimal movement in each configuration. Every two turns, the expert will play instead of the LA1 and perform the optimal action. We compare this with the performance of the LA1 when it solves the task alone, and with the one of a random policy.

\begin{figure}[thpb]
\centering
\includegraphics[scale=0.18]{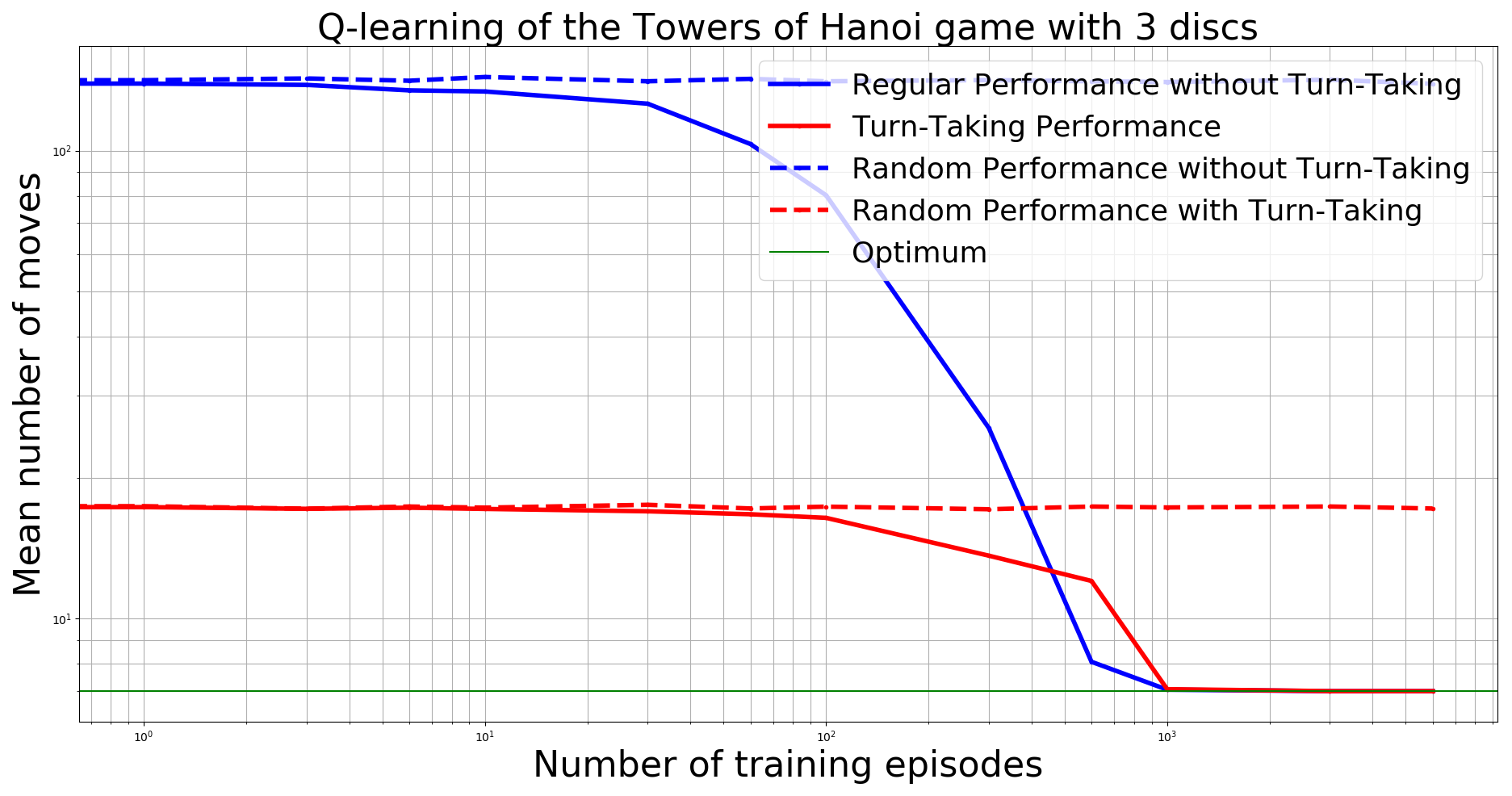}
\caption{Canonical intervention scenario: the LA1 solves the task in collaboration with the expert as they play alternatively. In cyan and yellow the performance when the LA1 follows a random policy, with and without help. Log. scale used on both axes.}
\label{fig1}
\end{figure}
   
As it can be seen in Fig. \ref{fig1}, the LA1 is directly more efficient when it is helped by the expert in a turn taking scenario, going from an order of $10^2$ moves to solve the task without help without training, to $10^1$ with canonical interventions. This can be explained by the fact that the agent is directly placed by the expert on the optimal sequence of actions (the left side diagonal from Fig. \ref{state}) to solve the task. In fact the expert is able to solve the task in 7 moves starting from any state, so after it has played, the LA1 is necessarily only 6 moves away from victory rather than 7. Thus during the first episodes of Q-learning, when the LA1 is not yet aware of the optimal path and acts somewhat randomly, it is still closer to the resolution of the task when it receives help than when it does not, because in the worst case it would be 6 moves away from the resolution instead of 7. In other words, the help of an expert improves the performance of the random policy. However the LA1 is moving away from the random policy after only 10 episodes when it does not receive help and it takes 100 training episodes to start drastically reducing its mean number of moves. At the same time, the performance still seems to be random in the turn-taking configuration and it takes to the agent 300 training episodes before it starts to converge to the optimal solution. The curves intersect after 400 training episodes when the LA1 without help starts to outperform the helped LA1. The LA1 needs 3,000 episodes of training to reach the optimal solution with canonical interventions, whereas it only needs 1,000 episodes when it is not helped. We can therefore conclude that being helped every 2 rounds by an expert agent does not speed up the learning process, on the contrary it slows it down.  

This is somehow not really surprising because the expert giving the optimal solution every two rounds prevents the agent from exploring every possible state. As shown in Fig. \ref{state}, the objective is to reach the \textit{222} state at the bottom left and each move of the expert will therefore lead the game to a state further to the left or further down than the previous state. This makes some states hard to reach (such as \textit{121}) or even impossible (such as those below the \textit{223}), thus delaying the convergence towards the optimal solution as the agent will still waste time trying to get in there even if it is not possible. This is a drawback of the learning system used. In contrast to some state-of-the-art methods such as Policy Shaping \cite{Griffith13,Cederborg15}, our Learning Agent is guided by an expert user and the feedback is not formulated as policy advice, as the goal is not to optimize 
 the human feedback but to mimic 
how a kid learns solve the task with a Learning Agent with a Q-learning algorithm, instead of a child as in the settings of \cite{Charisi20}.  The learning system could be improved by optimizing the teaching \cite{Cakmak12} by not always giving the optimal action but the one that will teach the agent the most.

A solution that would not deviate from the initial experimental setup could therefore be to let the LA1 explore the different states by involving the expert less frequently, by modifying the canonical intervention rate. This is what we did in Fig. \ref{fig2} in the Appendix, letting the expert play every 3 and 4 turns. 

\subsection{On demand or canonical intervention by the expert}

\begin{figure}[thpb]
\centering
\includegraphics[scale=0.18]{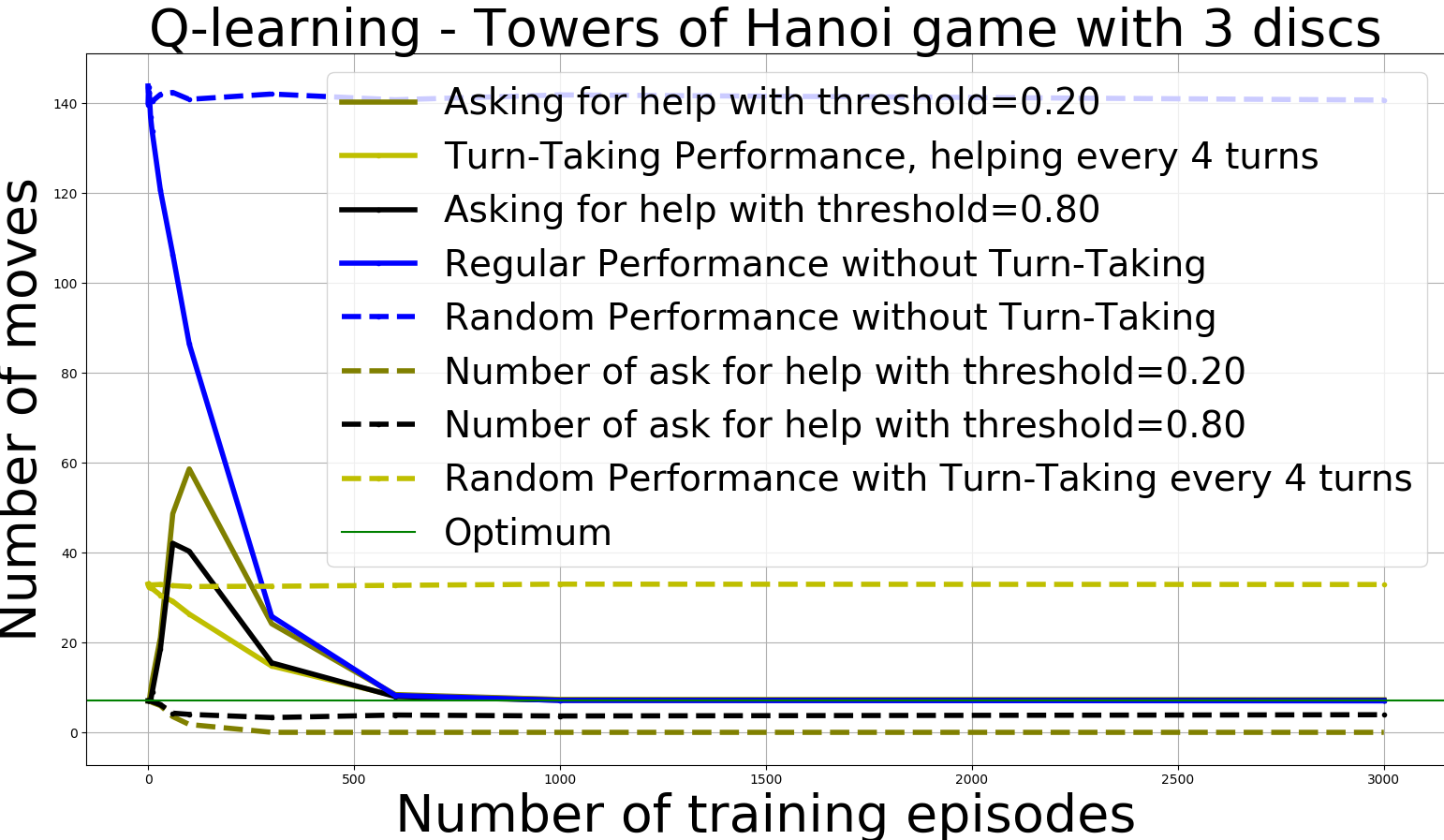}
\caption{Asking for help scenario with different \textit{ask for help} values: the LA2 tries to solve the game alone while being able to ask for help whenever its best action is not good enough (plot not on logarithmic scale as the agent asks for help at most 7 times).}
\label{fig:asking}
\end{figure}

The second configuration consists of a LA2, a Q-learning agent, trying to solve the task independently. It has the opportunity to ask for help to an expert agent whenever it needs to. To do this, we added an \textit{ask for help} parameter to the Q-learning. At each turn, if the best policy value is lower than the \textit{ask for help} parameter, the expert will play instead of the LA2. As we can see on Fig. \ref{fig:asking}, the LA2 is directly more effective, because he is always asking for help as it does not know yet what do to. After asking for help many times during the first 10 episodes it starts solving the task by itself, resulting in a loss of efficiency. We interpret this as the LA2 gaining confidence in movements which, while not perfect, still allows the task to progress towards its resolution through state exploration and trial. Compared to the LA1 without help, the LA2 asking for help is much more efficient but there is not a lot of variation between the canonical and the \textit{ask for help} configuration. This is probably due to the rather simple simulation of the ask for help trigger.

\section{DISCUSSION AND FUTURE WORK}

This paper presents the initial work towards the understanding of 
problem-solving process with two artificial agents by simulating a child-robot interaction experimental study. We acknowledge that the simulation of child's behaviour is a complex task and more emphasis is needed on accurate description of the multidimensional child behaviour.

The aspect of intrinsic motivation \cite{Oudeyer07}, indirectly related with solving a concrete task, but concerned with learning a set of reusable skills, should be further studied when rising the level of abstraction, specially in the context of solving different tasks and taking as input larger state spaces and of larger dimensionality \cite{Romero19, Prieto19} in order to simplify problem-solving in an end-to-end learning manner. State representation learning \cite{Lesort18} may come into use for a more realistic, less preprocessing demanding setting, i.e., not requiring human annotations of each game state when involving human collaboration.

Our point is to verify if hypotheses derived from an empirical study in an HRI setting are valid when translated from children to a RL Agent. Thus, the difficulty lies in the simulation of the child's behaviour by an artificial agent. The addition of an intrinsic motivation, on the desire for the LA to solve the game by itself, could increase the accuracy of this comparison. Our LA asks for help when it considers that a movement is not good enough to be played (i.e. when the largest Q-value among all available states fall under a pre-set threshold), whereas in reality, the mechanisms pushing the child to ask for help are much more complex \cite{Lake17, Biehl18}.


One of the challenges to explore is to validate the hypotheses tested with more complex tasks. More elaborated manners should be devised to more faithfully model uncertainty in the agent while acting. Future work could better mimic the presented and other human learning inspired behaviours. For instance, one could quantify (aleatoric and epistemic) uncertainty \cite{tagasovska2019single, Clements19,da2019uncertainty} of the agent's next action so we can better simulate the \textit{ask for help} setting when an agent is not certain enough. An accurate assessment should be made of the mechanisms that lead a child to ask for help when solving a task independently. This would allow it to be represented in the LA's behaviour so that it could ask for help in a more human-like natural way.

Future work includes the expansion of collaborative problem-solving settings with triadic interactions e.g. two children and a robot, in order to examine features of collective problem solving accounting for social dynamics \cite{Wallkotter20}. In addition to this, we are planning to examine the shifting processes \cite{miyake2000unity}, i.e. the processes of strategy generalization in a different task in human and artificial agents. 
Future work could also consider the possibility of trading-off between the gain generated for the agent by asking versus the disruption it causes to the human, using principles of mixed initiative interaction \cite{Horvitz99}. CoBots approaches\footnote{CoBots \url{http://www.cs.cmu.edu/~coral/projects/cobot/}} to ask for help are a related field to further explore, e.g., planning approaches for the LA to distinguish actions that it can complete autonomously from those that it cannot \cite{Rosenthal11,rosenthal2012someone}. 

Finally, in order to better understand child's developmental trajectories, we aim to replicate a similar child-robot interaction setting with a larger sample by manipulating additional variables such as the agent's social behaviour. This would inform our testing of more complex algorithms than Q-learning, using other dopamine based distributional RL signals \cite{dabney2020distributional}, and as little training data as people need \cite{Lake17}.

\section{ACKNOWLEDGEMENT}
We thank Cristina Conati for giving feedback on this work.

\bibliographystyle{abbrv} 
\bibliography{main}
\addtolength{\textheight}{-12cm}   



\section*{APPENDIX}

\subsection{Tower of Hanoï game} 
\begin{figure}[h!]
\centering
\includegraphics[scale=0.11]{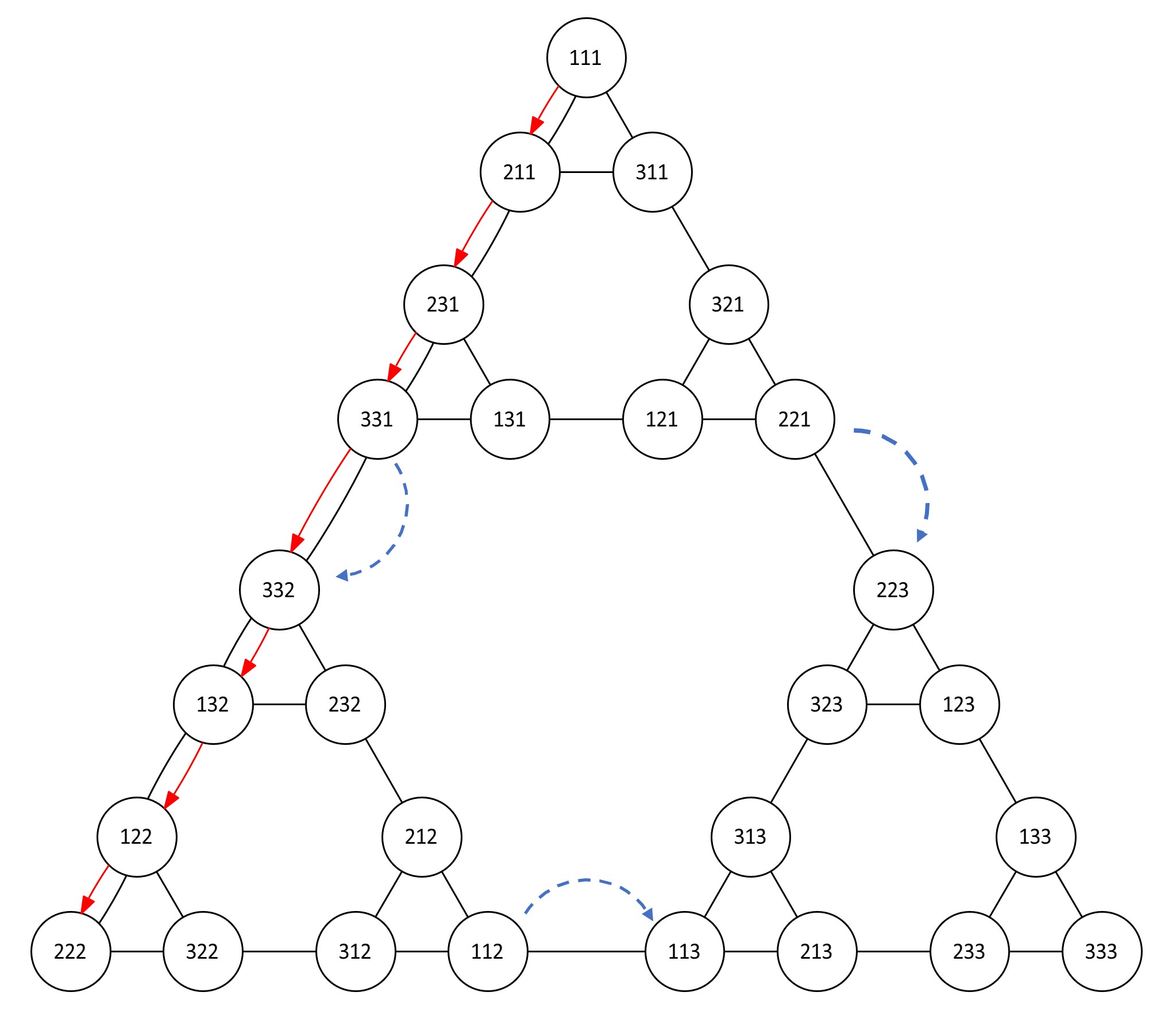} 
\caption{Abstract graph of the Tower of Hanoï game states for 3 disks. Each node represents one possible state of the game. The starting configuration is on the top, the final one in the bottom left. In red the optimal sequence of actions, in blue dashed movements between sub-graphs leading toward the solution (retrieved from \cite{Charisi20}.)}
\label{state}
\end{figure}

All possible states of the Tower of Hanoï game are in Fig. \ref{state}.

\subsection{Additional Results}

The LA1 who receives help is, regardless of the number of training episodes, always more efficient than the one who does not receive help. We can therefore conclude that an agent is more efficient when it receives help, as long as this help does not block its exploration.

\begin{figure}[htbp!]
\centering
\includegraphics[scale=0.18]{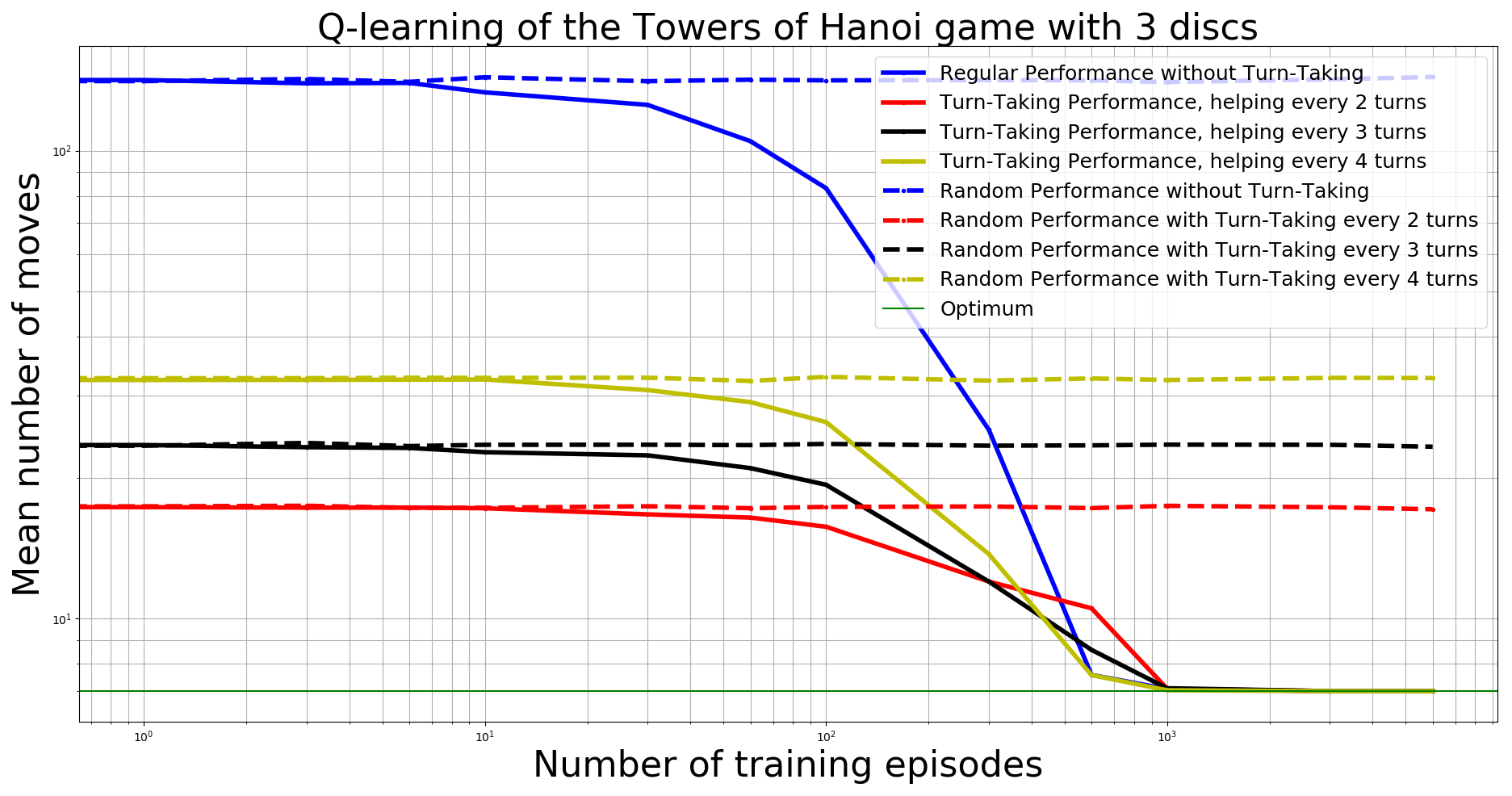}
\caption{Canonical intervention scenario with different intervention rates: the LA1 and the expert solve the game in collaboration but the expert is only playing every 2, 3 and 4 turns. It means that, e.g., in the last configuration, the LA1 will play 3 times before the expert plays. Log. scale used on both axes.}
\label{fig2}
\end{figure}





\end{document}